\documentclass[runningheads]{llncs}
\usepackage{graphicx}
\usepackage{amsmath,amssymb} 
\usepackage{color}
\usepackage{subfigure}
\usepackage[misc]{ifsym} 

\begin{document}
\title{Linear Span Network for Object Skeleton Detection} 

\titlerunning{Linear Span Network}

\author{Chang Liu\inst{*}  \and
Wei Ke\inst{*} \and
Fei Qin \and
Qixiang Ye \textsuperscript{(\Letter)}}
%
\authorrunning{C. Liu and W. Ke et al.}
%
\institute{University of Chinese Academy of Sciences, Beijing, China \\
\email{\{liuchang615, kewei11\}@mails.ucas.ac.cn, \{fqin1982, qxye\}@ucas.ac.cn } }
 \maketitle   
\begin{abstract}
Robust object skeleton detection requires to explore rich representative visual features and effective feature fusion strategies. In this paper, we first re-visit the implementation of HED, the essential principle of which can be ideally described with a linear reconstruction model. Hinted by this, we formalize a Linear Span framework, and propose Linear Span Network (LSN) which introduces Linear Span Units (LSUs) to minimizes the reconstruction error. LSN further utilizes subspace linear span besides the feature linear span to increase the independence of convolutional features and the efficiency of feature integration, which enhances the capability of fitting complex ground-truth. As a result, LSN can effectively suppress the cluttered backgrounds and reconstruct object skeletons. Experimental results validate the state-of-the-art performance of the proposed LSN.{\let\thefootnote\relax\footnotetext{$\ast$ indicates equal contributions.}}
{\let\thefootnote\relax\footnotetext{The source code is publicly available at https://github.com/LinearSpanNetwork.}}

\keywords{Linear Span Framework, Linear Span Unit, Linear Span Network, Skeleton Detection}
\end{abstract}
\section{Introduction}

Skeleton is one of the most representative visual properties, which describes objects with compact but informative curves. Such curves constitute a continuous decomposition of object shapes \cite{ref31}, providing valuable cues for both object representation and recognition. Object skeletons can be converted into descriptive features and spatial constraints, which enforce human pose estimation \cite{ref4}, semantic segmentation \cite{ref16}, and object localization \cite{ref32}. 

Researchers have been exploiting the representative CNNs for skeleton detection and extraction \cite{ref1,ref6,ref7,ref33} for years. State-of-the-art approaches root in effective multi-layer feature fusion, with the motivation that low-level features focus on detailed structures while high-level features are rich in semantics \cite{ref1}. As a pioneer work, the holistically-nested edge detection (HED) \cite{ref6} is computed as a pixel-wise classification problem, without considering the complementary among multi-layer features. Other state-of-the-art approaches, $e.g.$, fusing scale-associated deep side-outputs (FSDS) \cite{ref7,ref33} and side-output residual network (SRN) \cite{ref1} investigates the multi-layer association problem. FSDS requires intensive annotations of the scales for each skeleton point, while SRN struggles to pursuits the complementary between adjacent layers without complete mathematical explanation. The problem of how to principally explore and fuse more representative features remains to be further elaborated.

\begin{figure}[t]
\centering
\includegraphics[width=0.7\textwidth]{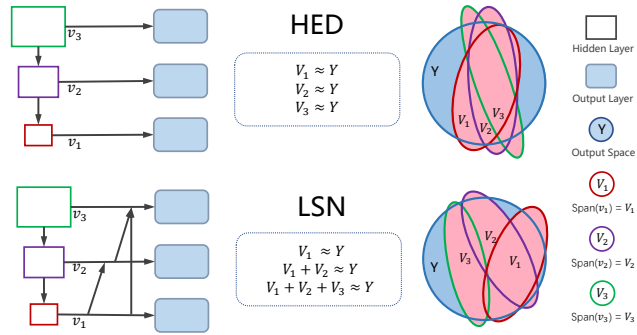}
\caption{A comparison of holistically-nested edge detection (HED) network \cite{ref6} and linear span network (LSN). HED uses convolutional features without considering their complementary. The union of the output spaces of HED is small, denoted as the pink area. As an improved solution, LSN spans a large output space. }
\label{fig:shot}
\end{figure}

Through the analysis, it is revealed that HED treats the skeleton detection as a pixel-wise classification problem with the side-output from convolutional network. Mathematically, this architecture can be equalized with a linear reconstruction model, by treating the convolutional feature maps as linear bases and the $1\times1$ convolutional kernel values as weights. Under the guidance of the linear span theory \cite{ref30}, we formalize a linear span framework for object skeleton detection. With this framework, the output spaces of HED could have intersections since it fails to optimize the subspace constrained by each other, Fig. 1. To ease this problem, we design Linear Span Unit (LSU) according to this framework, which will be utilized to modify convolutional network. The obtained network is named as Linear Span Network (LSN), which consists feature linear span, resolution alignment, and subspace linear span. This architecture will increase the independence of convolutional features and the efficiency of feature integration, which is shown as the smaller intersections and the larger union set, Fig. 1. Consequently, the capability of fitting complex ground-truth could be enhanced. By stacking multiple LSUs in a deep-to-shallow manner, LSN captures both rich object context and high-resolution details to suppress the cluttered backgrounds and reconstruct object skeletons.  
The contributions of the paper include: 
\begin{itemize}
     \item A linear span framework that reveals the essential nature of object skeleton detection problem, and proposes that the potential performance gain could be achieved with both the increased independence of spanning sets and the enlarged spanned output space.
     \item A Linear Span Network (LSN) can evolve toward the optimized architecture for 
     object skeleton detection under the guidance of linear span framework.    
\end{itemize}

\section{Related work}

Early skeleton extraction methods treat skeleton detection as morphological operations \cite{ref9,ref10,ref11,ref12,ref13,ref14,ref37}. One hypothesis is that object skeletons are the subsets of lines connecting center points of super-pixels \cite{ref12}. Such line subsets could be explored from super-pixels using a sequence of deformable discs to extract the skeleton path \cite{ref13}. In \cite{ref14}, The consistence and smoothness of skeleton are modeled with spatial filters, $e.g.$, a particle filter, which links local skeleton segments into continuous curves. Recently, learning based methods are utilized for skeleton detection. It is solved with a multiple instance learning approach \cite{ref15}, which picks up a true skeleton pixel from a bag of pixels. The structured random forest is employed to capture diversity of skeleton patterns \cite{ref16}, which can be also modeled with a subspace multiple instance learning method \cite{ref17}.

With the rise of deep learning, researchers have recently formulated skeleton detection as image-to-mask classification problem by using learned weights to fuse the multi-layer convolutional features in an end-to-end manner. HED \cite{ref6} learns a pixel-wise classifier to produce edges, which can be also used for skeleton detection. Fusing scale-associated deep side-outputs (FSDS) \cite{ref7} learns multi-scale skeleton representation given scale-associated ground-truth. Side-output residual network (SRN) \cite{ref1} leverages the output residual units to fit the errors between the object symmetry/skeleton ground-truth and the side-outputs of multiple convolutional layers. 

The problem about how to fuse multi-layer convolutional features to generate an output mask, $e.g.$, object skeleton, has been extensively explored. Nevertheless, existing approaches barely investigate the problem about the linear independence of multi-layer features, which limits their representative capacity. Our approach targets at exploring this problem from the perspective of linear span theory by feature linear span of multi-layer features and subspace linear span of the spanned subspaces.

\section{Problem Formulation}

\subsection{Re-thinking HED}
\begin{figure}[t]
\centering
\subfigure[dependent linear span]{\includegraphics[width=1.7in]{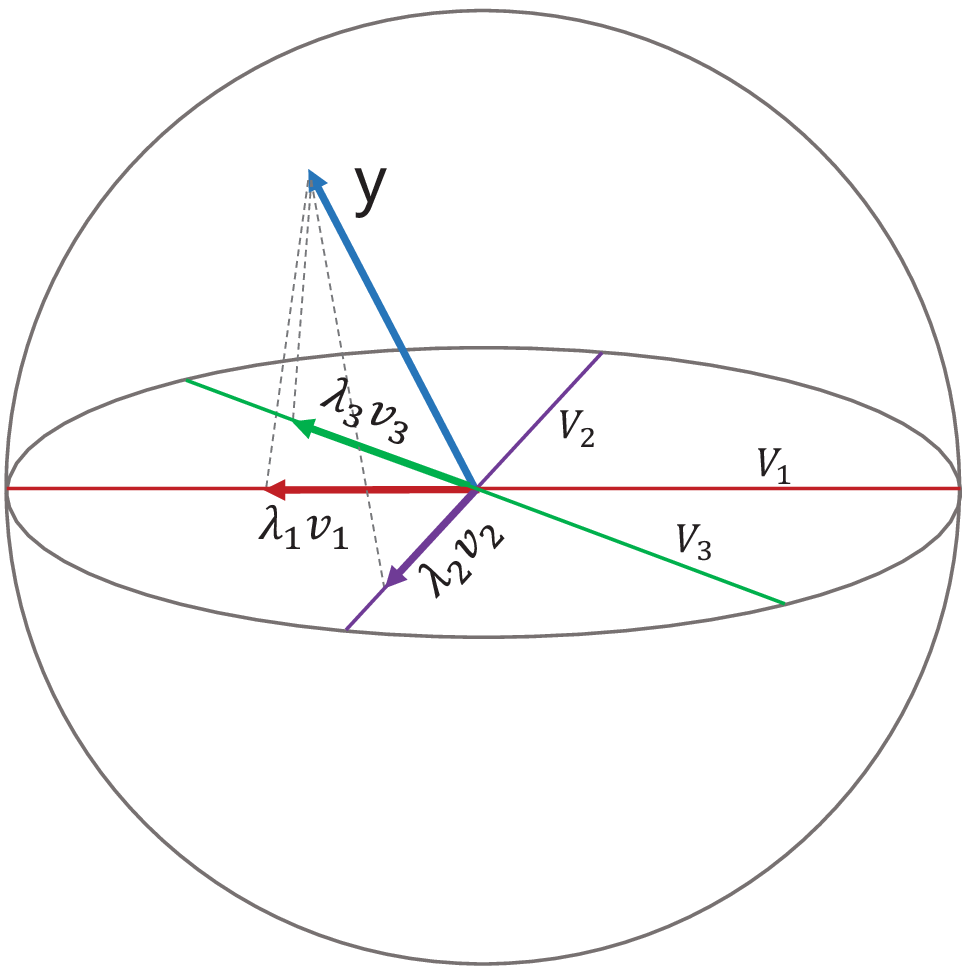}}
\qquad\quad 
\subfigure[independent linear span]{\includegraphics[width=1.7in]{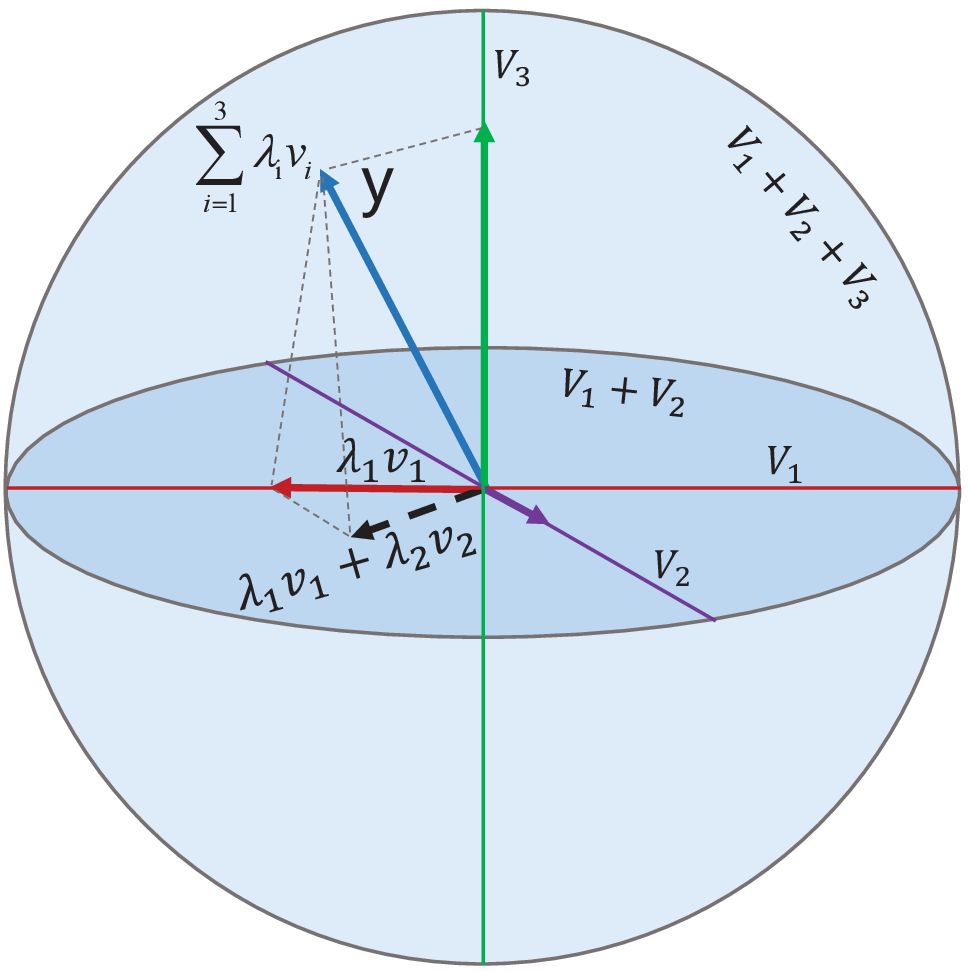}} 
\caption{Schematic of linear span with a set of dependent vectors (a) and independent vectors (b).  }
\label{fig2:vector}
\end{figure}

In this paper, we re-visit the implementation of HED, and reveal  that HED as well as its variations can be all formulated by the linear span theory \cite{ref30}.

HED utilizes fully convolutional network with deep supervision for edge detection, which is one of the typical low-level image-to-mask task. Denoting the convolutional feature as $C$ with $m$ maps and the classifier as $w$, HED is computed as a pixel-wise classification problem, as
\begin{equation}
   {\hat {y_j}} = \sum\limits_{k = 1}^m {{w_k} \cdot {c_{k,j}}} ,j = 1,2, \cdots ,|{\hat Y}|,
   \label{hed::classifier}
\end{equation}
where ${c_{k,j}}$ is the feature value of the $j$-th pixel on the $k$-th convolutional map and ${\hat y_j}$ is the classified label of the $j$-th pixel in the output image ${\hat Y}$.

Not surprisingly, this can be equalized as a linear reconstruction problem, as
\begin{equation}
   { Y} = \sum\limits_{k = 1}^m {{\lambda _k}{v_k}},
   \label{hed::reconstruction}
\end{equation}
where ${\lambda _k}$ is linear reconstruction weight and $v_k$  is the $k$-th feature map in $C$. 

We treat each side-output of HED as a feature vector in the linear spanned subspace $V_i=span({v_1^i,v_2^i,\cdots,v_m^i})$, in which $i$ is the index of convolutional stages. Then HED forces each subspace $V_i$ to approximate the ground-truth space $\mathcal{Y}$. We use three convolutional layers as an example, which generate subspaces $V_1$, $V_2$, and $V_3$. Then the relationship between the subspaces and the ground-truth space can be illustrated as lines in a 3-dimension space in Fig. 2(a).  

As HED does not optimize the subspaces constrained by each other, it fails to explore the complementary of each subspace to make them decorrelated. The reconstructions can be formulated as
\begin{equation}
\left\{\begin{array}{l}
V_1 \approx \mathcal{Y}\\ 
V_2 \approx \mathcal{Y}\\
V_3 \approx \mathcal{Y}
\end{array}.\right.
\label{hed:span}
\end{equation}
When $v_1$, $v_2$, and $v_3$ are linearly dependent, they only have the capability to reconstruct vectors in a plane. That is to say, when the point $Y$ is out of the plane, the reconstruction error is hardly eliminated, Fig. 2(a). 

Obviously, if $v_1$, $v_2$, and $v_3$ are linearly independent, $i.e.$, not in the same plane, Fig. 2(b), the reconstruction could be significantly eased. To achieve this target, we can iteratively formulate the reconstruction as
\begin{equation}
\left\{\begin{array}{l}
V_1 \approx \mathcal{Y}\\ 
V_1 + V_2 \approx \mathcal{Y} \\
V_1 + V_2 + V_3 \approx \mathcal{Y} 
\end{array}.\right.
\label{lsn::span}
\end{equation}
It s observed that $V_2$ is refined with the constraint of $V_1$. And $V_3$ is optimized in the similar way, which aims for vector decorrelation.  The sum of subspaces, $i.e.$, $V_1+V_2$ is denoted with the dark blue plane, and $V_1+V_2+V_3$ is denoted with the light blue sphere, Fig. 2(b). 

Now, it is very straightforward to generalize Eq. (\ref{lsn::span}) to 
\begin{equation}
   \sum\limits_{k = 1}^l {{V_k}} \approx \mathcal{Y}, l=1,2,\cdots,n.
   \label{genera::span}
\end{equation}
One of the variations of HED, $i.e.$, SRN, which can be understand as a special case of Eq. (5) with $\sum\nolimits_{k = l - 1}^l {{V_k}} \approx \mathcal{Y}$, has already shown the effectiveness.

\subsection{Linear Span View}

Based on the discussion of last section, we can now strictly formulate a mathematical framework based on linear span theory \cite{ref30}, which can be utilized to guide the design of Linear Span Network (LSN) toward the optimized architecture. 

In linear algebra, linear span is defined as a procedure to construct a linear space by a set of vectors or a set of subspaces.

\textbf{\textit{Definition 1.}} $\mathcal{Y}$ is a linear space over ${ \mathbb{R}}$. The set $\left\{ {{v_1},{v_2},...,{v_m}} \right\} \subset \mathcal{Y}$ is a spanning set for $\mathcal{Y}$ if every $y$ in $\mathcal{Y}$ can be expressed as a linear combination of ${v_1},{v_2},...,{v_m}$, as
\begin{equation}
    y = \sum\limits_{k=1}^{m} {{\lambda_k}{v_k}}, \ \lambda_1,...,\lambda_m \in \mathbb{R},
    \label{ls:def1}
\end{equation}
and $\mathcal{Y}=span(\left\{ {{v_1},{v_2},...,{v_m}}\right\})$.

\textbf{\textit{Theorem 1.} } Let ${v_1},{v_2},...,{v_m}$ be vectors in $\mathcal{Y}$. Then $\left\{ {{v_1},{v_2},...,{v_m}} \right\}$ spans $\mathcal{Y}$ if and only if, for the matrix $F = \left[ {{v_1}{\rm{  }}{v_2}{\rm{  }}...{\rm{  }}{v_m}} \right]$, the linear system $F\lambda = y$ is consistent for every $y$ in $\mathcal{Y}$. 

\textbf{\textit{Remark 1.} } According to \textit{Theorem} 1, if the linear system is consistent for almost every vector in a linear space, the space can be approximated by the linear spanned space. This theorem uncovers the principle of LSN, which pursues a linear system as mentioned above setting up for as many as ground-truth.

\textbf{\textit{Definition 2.} } A finite set of vectors, which span  $\mathcal{Y}$ and are linearly independent, is called a basis for $\mathcal{Y}$.

\textbf{\textit{Theorem 2. }} Every linearly independent set of vectors $\left\{ {{v_1},{v_2},...,{v_m}} \right\}$ in a finite dimensional linear space $\mathcal{Y}$ can be completed to a basis of $\mathcal{Y}$.

\textbf{\textit{Theorem 3.}}  Every subspace $U$ has a complement in $\mathcal{Y}$, that is, another subspace $V$ such that vector $y$ in $\mathcal{Y}$ can be decomposed uniquely as 
\begin{equation}
   y = u + v, u \ in \ U, v \ in \ V.
   \label{ls::theo3}
\end{equation}	  

\textbf{\textit{Definition 3.} } $\mathcal{Y}$ is said to be the sum of its subspaces $V_1,...,V_m$ if every $y$ in $Y$ can be expressed as
\begin{equation}
    y = v_1+...+v_m, v_j \  in \  V_j.
    \label{ls::def3}
\end{equation}

\textbf{\textit{Remark 2.}} We call the spanning of feature maps to a subspace as feature linear span, and the sum of subspaces as subspace linear span. From \textit{Theorem 2} and \textit{Theorem} 3, it is declared that the union of the spanning sets of subspaces is the spanning set of the sum of the subspaces. That is to say, in the subspace linear span we can merge the spanning sets of subspaces step by step to construct a larger space.

\textbf{\textit{Theorem 4.}} Supposing $\mathcal{Y}$ is a finite dimensional linear space, $U$ and $V$ are two subspaces of $\mathcal{Y}$ such that $\mathcal{Y} = U + V $, and $W$ is the intersection of $U$ and $V$, $i.e.$, $W = U \cap V $. 
Then
\begin{equation}
   \dim \mathcal{Y} = \dim U + \dim V - \dim W.
\end{equation}	   

\textbf{\textit{Remark 3.}} From \textit{Theorem} 4, the smaller the dimension of the intersection of two subspaces is, the bigger the dimension of the sum of two subspaces is. Then, successively spanning the subspaces from deep to shallow with supervision increases independence of spanning sets and enlarges the sum of subspaces. It enfores the representative capacity of convolutional features and integrates them in a more effective way.

\section{Linear Span Network}

With the help of the proposed framework, the Linear Span Network (LSN) is designed for the same targets with HED and SRN, $i.e.$, the object skeleton detection problem. Following the linear reconstruction theory, a novel architecture named Linear Span Unit(LSU) has been defined first. Then, LSN is updated from VGG-16 \cite{ref33} with LSU and hints from \textit{Remark} 1-3. VGG-16 has been chosen for the purpose of fair comparison with HED and SRN. In what follows, the implementation of LSU and LSN are introduced.

\subsection{Linear Span Unit}

\begin{figure}[t]
\centering
\includegraphics[width=0.8\textwidth]{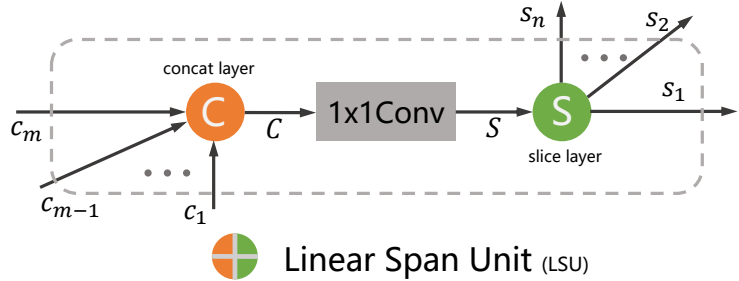}
\caption{Linear Span Unit, which is used in both feature linear span and subspace linear span. In LSU, the operation of linear reconstruction is implemented by a concatenation layer and a $1\times1$ convolutional layer. }
\label{fig:lsu}
\end{figure}

The architecture of Linear Span Unit (LSU) is shown in Fig. \ref{fig:lsu}, where each feature map is regarded as a feature vector. The input feature vectors are unified with a concatenation (concat for short) operation, as 

\begin{equation}
   {C} = \mathop {concat}\limits_{k = 1}^{m} {\rm{ }}(c_{k}),
   \label{lsu::concat}
\end{equation}
where $c_k$ is the $k$-th feature vector. In order to compute the linear combination of the feature vectors, a convolution operation with $1\times1\times m$ convolutional kernels is employed:
\begin{equation}
   {{s_i}} = \sum\limits_{k = 1}^m {{\lambda_{k,i}} \cdot {c_{k}}} ,i = 1,2, \cdots ,n,
   \label{lsu::reconstructor}
\end{equation}
where ${\lambda_{k,i}}$ is the convolutional parameter with $k$ elements for the $i$-th reconstruction output. The LSU will generate $n$ feature vectors in the subspace spanned by the input feature vectors. A slice layer is further utilized to separate them for different connections, which is denoted as
\begin{equation}
    \bigcup\limits_{i=1}^{n} s_i = slice({{\rm{S}}}).
    \label{lsu::slice}
\end{equation}

\subsection{Linear Span Network Architecture}
\begin{figure}[t]
\centering
\includegraphics[width=0.98\textwidth]{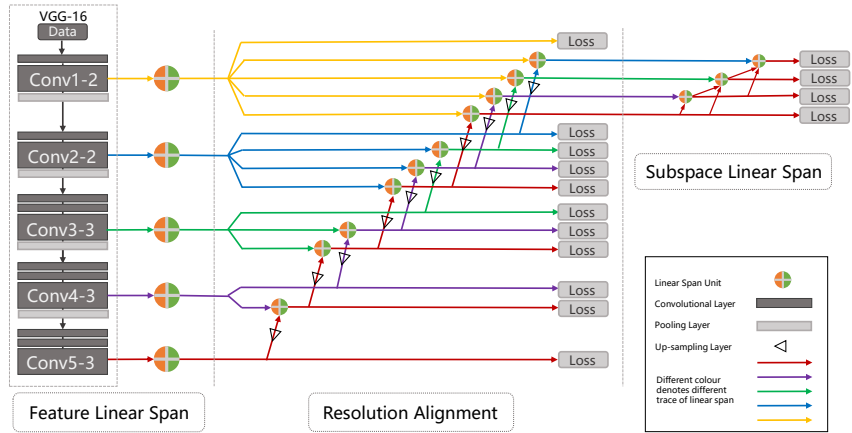}
\caption{The architecture of the proposed Linear Span Network (LSN), which leverages Linear Span Units (LSUs) to implement three components of the feature linear span, the resolution alignment, and the subspace linear span. 
The feature linear span uses convolutional features to build subspaces. The LSU is re-used to unify the resolution among multi-stages in resolution alignment. The subspace linear span summarizes the subspaces to fit the ground-truth space. }
\label{fig:LSN}
\end{figure}

The architecture of LSN is shown in Fig.\ \ref{fig:LSN}, which is consisted of three components, $i.e.$, feature linear span, resolution alignment, and subspace linear span are illustrated. The VGG-16 network with 5 convolutional stages \cite{ref34} is used as the backbone network. 

In feature linear span, LSU is used to span the convolutional feature of the last layer of each stage according to Eq. \ref{lsu::reconstructor}. The supervision is added to the output of LSU so that the spanned subspace approximates the ground-truth space, following \textit{Remark} 1. If only feature linear span is utilized, the LSN is degraded to HED \cite{ref6}. Nevertheless, the subspaces in HED separately fit the ground-truth space, and thus fail to decorrelate spanning sets among subspaces. According to \textit{Remark} 2 and 3, we propose to further employ subspace linear span to enlarge the sum of subspaces and deal with the decorrelation problem. 

As the resolution of the vectors in different subspaces is with large variation, simple up-sampling operation will cause the Mosaic effect, which generates noise in subspace linear span. Without any doubt, the resolution alignment is necessary for LSN. Thus, in Fig.~\ref{fig:LSN}, LSUs have been laid between any two adjacent layers with supervision. As a pre-processing component to subspace linear span, it outputs feature vectors with same resolution.

The subspace linear span is also implemented by LSUs, which further concatenates feature vectors from deep to shallow layers and spans the subspaces with Eq. (\ref{genera::span}). According to \textit{Remark} 3, a step-by-step strategy is utilized to explore the complementary of subspaces. With the loss layers attached on LSUs, it not only enlarges the sum of subspaces spanned by different convolutional layers, but also decorrelates the union of spanning sets of different subspaces. With this architecture, LSN enforces the representative capacity of convolutional features to fit complex ground-truth.

\section{Experiments}

\subsection{Experimental setting}
\textbf{Datasets:} We evaluate the proposed LSN on pubic skeleton datasets including SYMMAX \cite{ref15}, WH-SYMMAX \cite{ref17}, SK-SMALL \cite{ref7}, SK-LARGE \cite{ref33}, and Sym-PASCAL\cite{ref1}. We also evaluate LSN to edge detection on the BSDS500 dataset \cite{ref35} to validate its generality.

SYMMAX is derived from BSDS300 \cite{ref35}, which contains 200/100 training and testing images. It is annotated with local skeleton on both foreground and background. WH-SYMMAX is developed for object skeleton detection, but contains only cropped horse images, which are not comprehensive for general object skeleton. SK-SMALL involves skeletons about 16 classes of objects with 300/206 training and testing images. Based on SK-SMALL, SK-LARGE is extended to 746/745 training and testing images. Sym-PASCAL is derived from the PASCAL-VOC-2011 segmentation dataset \cite{ref36} which contains 14 object classes with 648/787 images for training and testing. 

The BSDS500 \cite{ref35} dataset is used to evaluate LSN's performance on edge detection. This dataset is composed of 200 training images, 100 validation images, and 200 testing images. Each image is manually annotated by five persons on average. For training images, we preserve their positive labels annotated by at least three human annotators.

\textbf{Evaluation protocol:} Precision recall curve (PR-curve) is use to evaluate the performance of the detection methods. With different threshold values, the output skeleton/edge masks are binarized. By comparing the masks with the ground-truth, the precision and recall are computed. For skeleton detection, the F-measure is used to evaluate the performance of the different detection approaches, which is achieved with the optimal threshold values over the whole dataset, as
\begin{equation}
   F = \frac{{2PR}}{{P + R}}.
\end{equation}
To evaluate edge detection performance, we utilize three standard measures \cite{ref35}: F-measures when choosing an optimal scale for the entire dataset (ODS) or per image (OIS), and the average precision (AP).

\textbf{Hyper-parameters:} For both skeleton and edge detection, we use VGG16 \cite{ref34} as the backbone network. During learning we set the mini-batch size to 1, the loss-weight  to 1 for each output layer, the momentum to 0.9, the weight decay to 0.002, and the initial learning rate to 1e-6, which decreases one magnitude for every 10,000 iterations. 

\subsection{LSN Implementation}

We evaluate four LSN architectures for subspace linear span and validate the iterative training strategy. 

\textbf{LSN architectures.}
If there is no subspace linear span, Fig.\ \ref{fig:LSN}, LSN is simplified to HED \cite{ref6}, which is denoted as LSN\_1. The F-measure of LSN\_1 is 49.53\%. When the adjacent two subspaces are spanned, it is denoted as LSN\_2, which is the same as SRN \cite{ref1}. LSN\_2 achieve significant performance improvement over HED which has feature linear span but no subspace span. We compare LSNs with different number of subspaces to be spanned, and achieve the best F-measure of 66.82\%. When the subspace number is increased to 4, the skeleton detection performance drops. The followings explained why LSN\_3 is the best choice.

If the subspaces to be spanned are not enough, the complementary of convolutional features from different layers could not be effectively explored. On the contrary, if a LSU fuses feature layers that have large-scale resolution difference, it requires to use multiple up-sampling operations, which deteriorate the features. Although resolution alignment significantly eases the problem, the number of adjacent feature layers to be fused in LSU remains a practical choice. LSN\_3 reported the best performance by fusing a adjacent layer of higher resolution and a adjacent layer of lower resolution.On one hand, the group of subspaces in LSN\_3 uses more feature integration. On the other hand, there is not so much information loss after an $2\times$ up-sampling operation.

\setlength{\tabcolsep}{20pt}
\begin{table}[t]
\begin{center}
\caption{The performance of different LSN implementations on the SK-LARGE dataset. LSN\_3 that fuses an adjacent layer of higher resolution and an adjacent layer of lower resolution reported the best performance. }
\begin{tabular}{lc}
\hline\noalign{\smallskip}
Architecture & F-measure(\%) \\
\noalign{\smallskip}
\hline
\noalign{\smallskip}
LSN\_1 (HED, feature linear span only)& 49.53 \\
LSN\_2 (SRN, feature and 2-subspace linear span)&65.88 \\
LSN\_3 (LSN, feature and 3-subspace linear span)&\textbf{66.15}\\
LSN\_4 (LSN, feature and 4-subspace linear span)&65.89 \\
\hline
\label{LSU_Table}
\end{tabular}
\end{center}
\end{table}
\setlength{\tabcolsep}{1.4pt}

\setlength{\tabcolsep}{8pt}
\begin{table}[h]
\begin{center}
\caption{The performance for different training strategies.}
\begin{tabular}{lccccc}
\hline\noalign{\smallskip}
   &w/o RA & end-to-end & iter1 & iter2 & iter3 \\
\noalign{\smallskip}
\hline
\noalign{\smallskip}
F-measure(\%)&66.15 & 66.63 & \textbf{66.82} & 66.74 & 66.68 \\
\hline
\label{Training_strategy_table}
\end{tabular}
\end{center}
\end{table}
\setlength{\tabcolsep}{1.4pt}

\textbf{Training strategy.}
With three feature layers spanned, LSN needs up-sampling the side-output feature layers from the deepest to the shallowest ones. We use the supervised up-sampling to unify the resolution of feature layers.

During training, the resolution alignment is also achieved by stacking LSUs.
We propose a strategy that train the two kinds of linear span, $i.e.$, feature linear span with resolution alignment and subspace linear span, iteratively. In the first iteration, we tune the LSU parameters for feature linear span and resolution alignment using the pre-trained VGG model on ImageNet, as well as update the convolutional parameters. Keeping the LSU parameters for resolution alignment unchanged, we tune LSU parameters for feature linear span and subspace linear span using the new model. In other iteration, the model is fine-tuned on the snap-shot of the previous iteration. With this training strategy, the skeleton detection performance is improved from 66.15\% to 66.82\%, Table\ \ref{Training_strategy_table}. The detection performance changes marginally when more iterations are used. We therefore use the single iteration (iter1) in all experiments.

\begin{figure}[t]
\centering
\subfigure[Feature linear span by HED]{\includegraphics[width=2.3in]{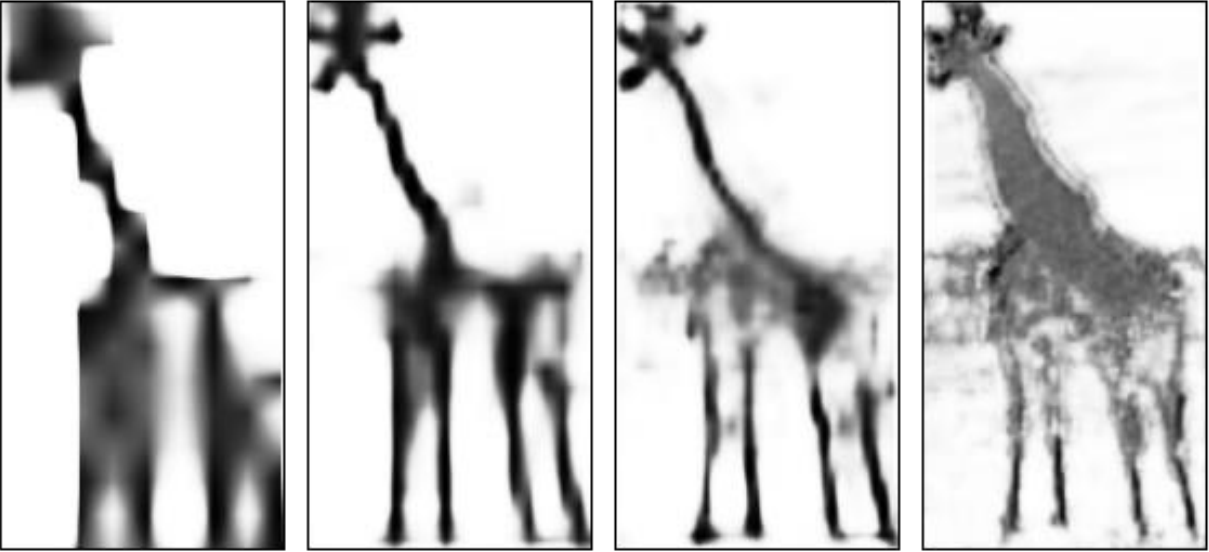}}
\quad 
\subfigure[Subspace linear span by SRN]{\includegraphics[width=2.3in]{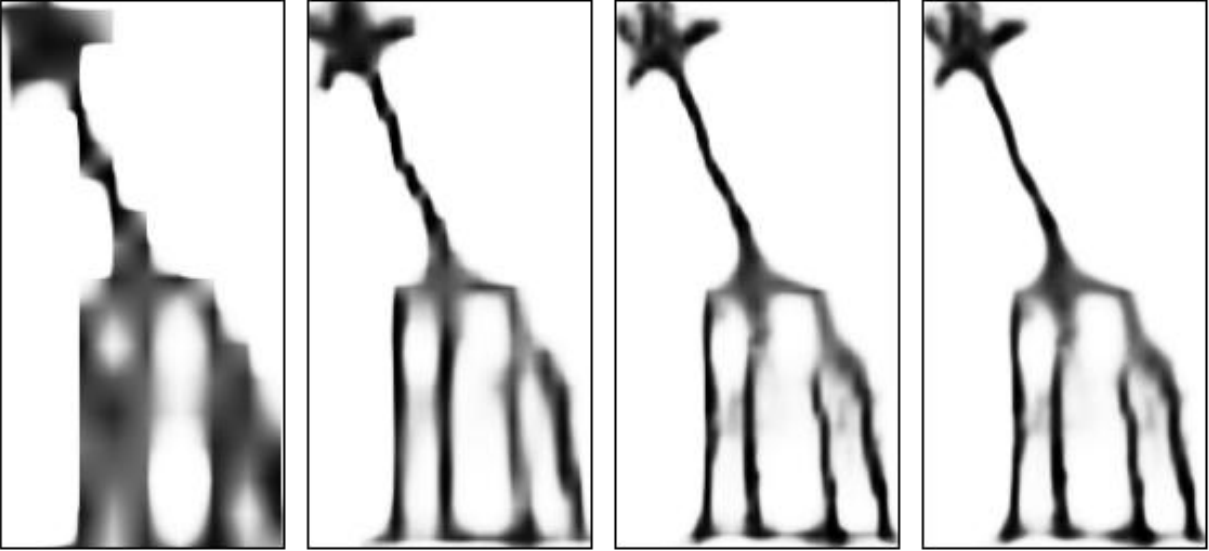}} \\
\subfigure[Feature linear span by LSN]{\includegraphics[width=2.3in]{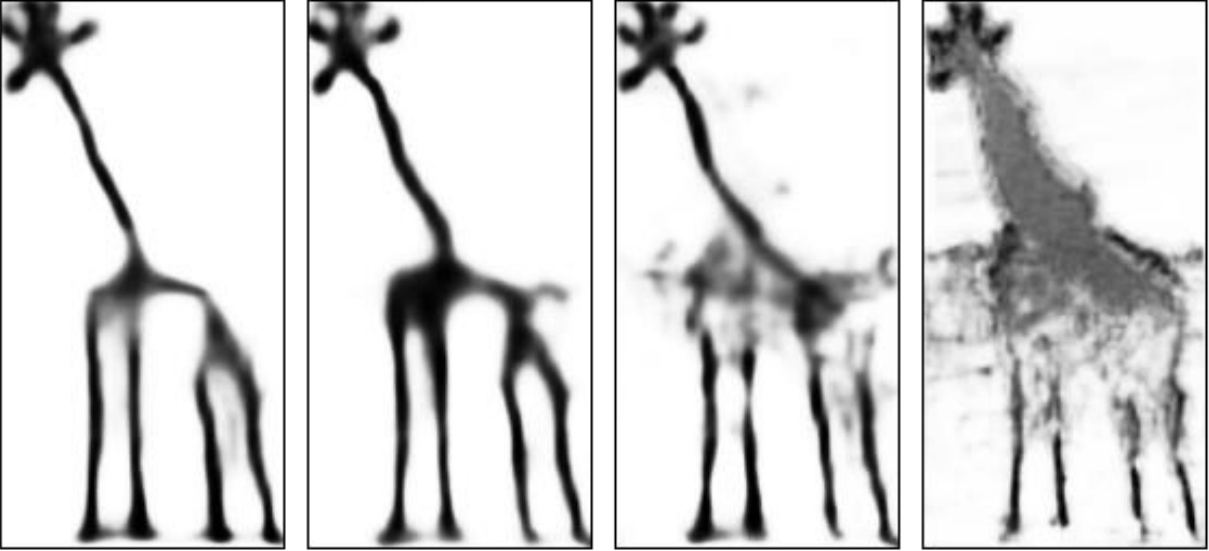}}
\quad 
\subfigure[Subspace linear span by LSN]{\includegraphics[width=2.3in]{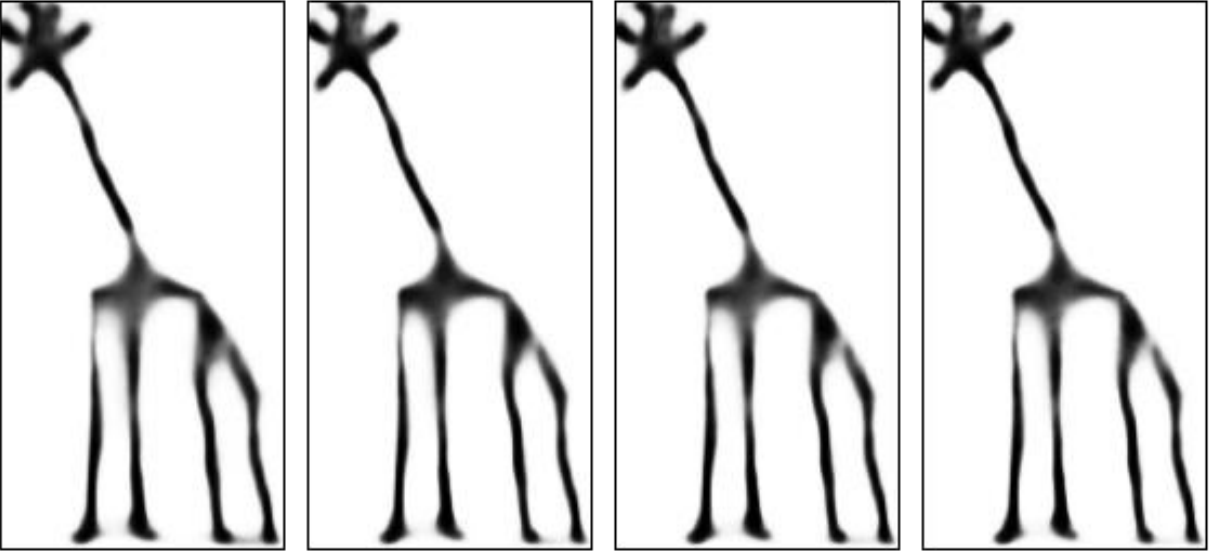}}
\caption{Comparison of output feature vectors of HED \cite{ref6}, SRN \cite{ref1}, and LSN(From left to right results are listed in a deep-to-shallow manner). By comparing (a) and (c), (b) and (d), one can see that LSN can learn better feature vectors and subspaces(basis) to span the output space. It enforces the representative capacity of convolutional features to fit complex outputs with limited convolutional layers.
}
\label{Bases}
\end{figure}

\textbf{LSU effect.}
In Fig.\ \ref{Bases}, we use a giraffe's skeleton from SK-LARGE as an example to compare and analyze the learned feature vectors (bases) by HED \cite{ref6}, SRN \cite{ref6}, and LSN. In Fig.\ \ref{Bases}(a) and (c), we respectively visualize the feature vectors learned by HED \cite{ref6} and the proposed LSN. It can be seen in the first column that the HED's results incorporate more background noise and mosaic effects. This shows that the proposed LSN can better span an output feature space. In Fig.\ \ref{Bases}(b) and (d), we respectively visualize the subspace vectors learned by SRN \cite{ref1} and the proposed LSN. It can be seen in the first column that the SRN's results incorporate more background noises. It requires to depress such noises by using a residual reconstruction procedure. In contrast, the subspace vectors of LSN is much clearer and compacter. This fully demonstrates that LSN can better span the output space and enforce the representative capacity of convolutional features, which will ease the problems of fitting complex outputs with limited convolutional layers. 

\subsection{Performance and Comparison}

{
\setlength\parindent{-0.5em}
\begin{minipage}{.48\textwidth}
\centering
\includegraphics[width=2.3in]{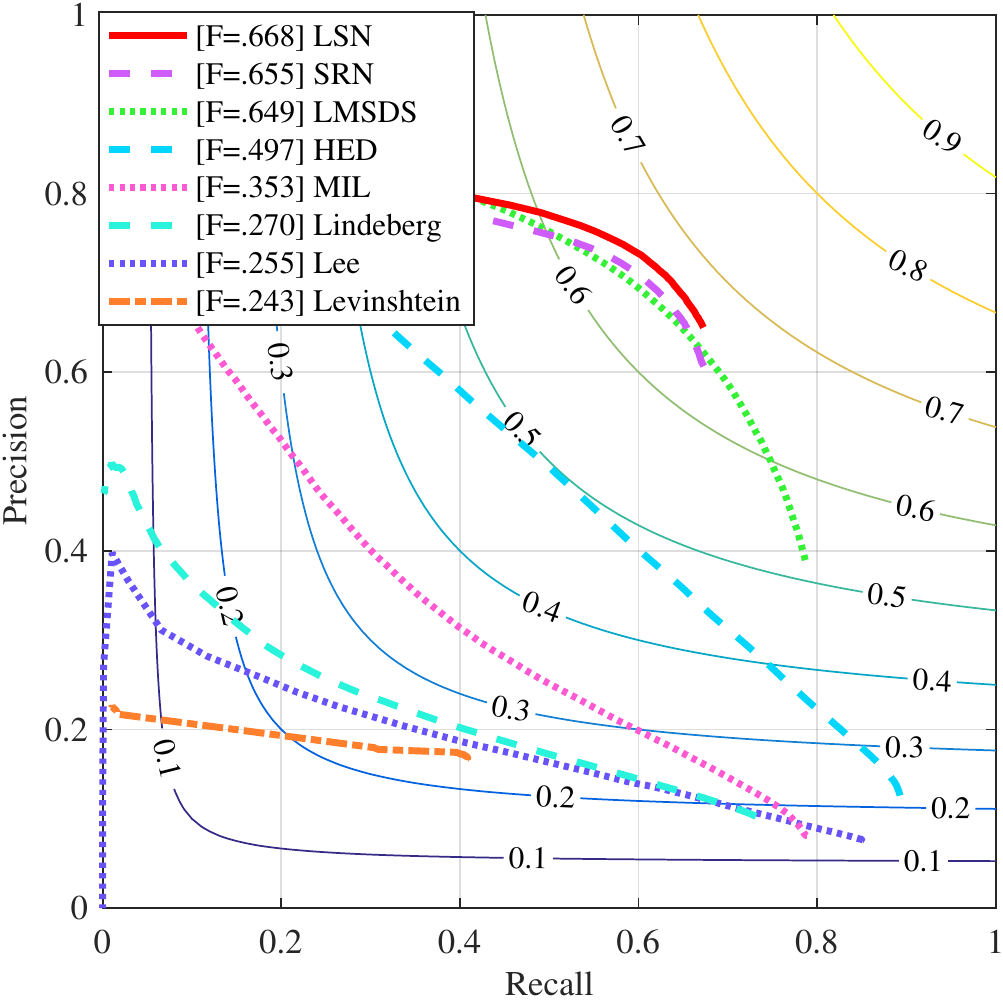}
\makeatletter\def\@captype{figure}\makeatother
\caption{The PR-curve on SK-LARGE.  }
\end{minipage} \quad
\begin{minipage}{.49\textwidth}
  \makeatletter\def\@captype{table}\makeatother
  \caption{Performance comparison  on SK-LARGE dataset. $\dagger$ GPU time.}
  \begin{tabular}{lcc}
    \hline
    Mehods & F-measure & Runtime/s \\
    \noalign{\smallskip}
    \hline
    \noalign{\smallskip}
    Lindeberg \cite{ref37} &	0.270 &	4.05 \\
    Levinshtein \cite{ref12} &	0.243 &	146.21 \\
    Lee \cite{ref13} &	0.255 &	609.10 \\
    MIL \cite{ref15} &	0.293 &	42.40 \\
    \hline
    HED \cite{ref6} &	0.495 &	\textbf{0.05} $\dagger$ \\
    SRN \cite{ref1} &	0.655 &	0.08 $\dagger$ \\
    LMSDS \cite{ref33} &	0.649 &	\textbf{0.05} $\dagger$ \\
    LSN (ours) & \textbf{0.668} &	0.09 $\dagger$ \\
    \hline
    \label{tab:sk_large}
  \end{tabular}
  \label{fig:sklarge}
\end{minipage}
}

\textbf{Skeleton detection.} 

The proposed LSN is evaluated and compared with the state-of-the-art approaches, and the performance is shown in Fig. \ref{fig:sklarge} and Table \ref{tab:sk_large}. The result of SRN \cite{ref1} is obtained by running authors' source code on a Tesla K80 GPU, and the other results are provided by \cite{ref33}.

The conventional approaches including Lindeberg \cite{ref37}, Levinshtein \cite{ref12}, and Lee \cite{ref13}, produce the skeleton masks without using any learning strategy. They are time consuming and achieve very low F-measure of 27.0\%, 24.3\%, and 25.5\%, respectively. The typical learning approach, $i.e.$, multiple instance learning (MIL) \cite{ref15}, achieves F-measure of 29.3\%. It extractes pixel-wised feature with multi-orientation and multi-scale, and averagely uses 42.40 seconds to distinguish skeleton pixels from the backgrounds in a single image.

The CNN based approaches achieve huge performance gain compared with the conventional approaches. HED \cite{ref6} achieves the F-measure of 49.5\% and uses  0.05 seconds to process an images, while SRN \cite{ref1} achieves 64.9\% and uses 0.08 seconds. The scale-associated multi-task method, LMSDS \cite{ref33}, achieves the performance of 64.9\%, which is built on HED with the pixel-level scale annotations. Our proposed LSN reportes the best detection performance of 66.8\% with a little more runtime cost compared with HED and SRN. 

\begin{figure}[h]
\centering
\includegraphics[width=0.95\linewidth]{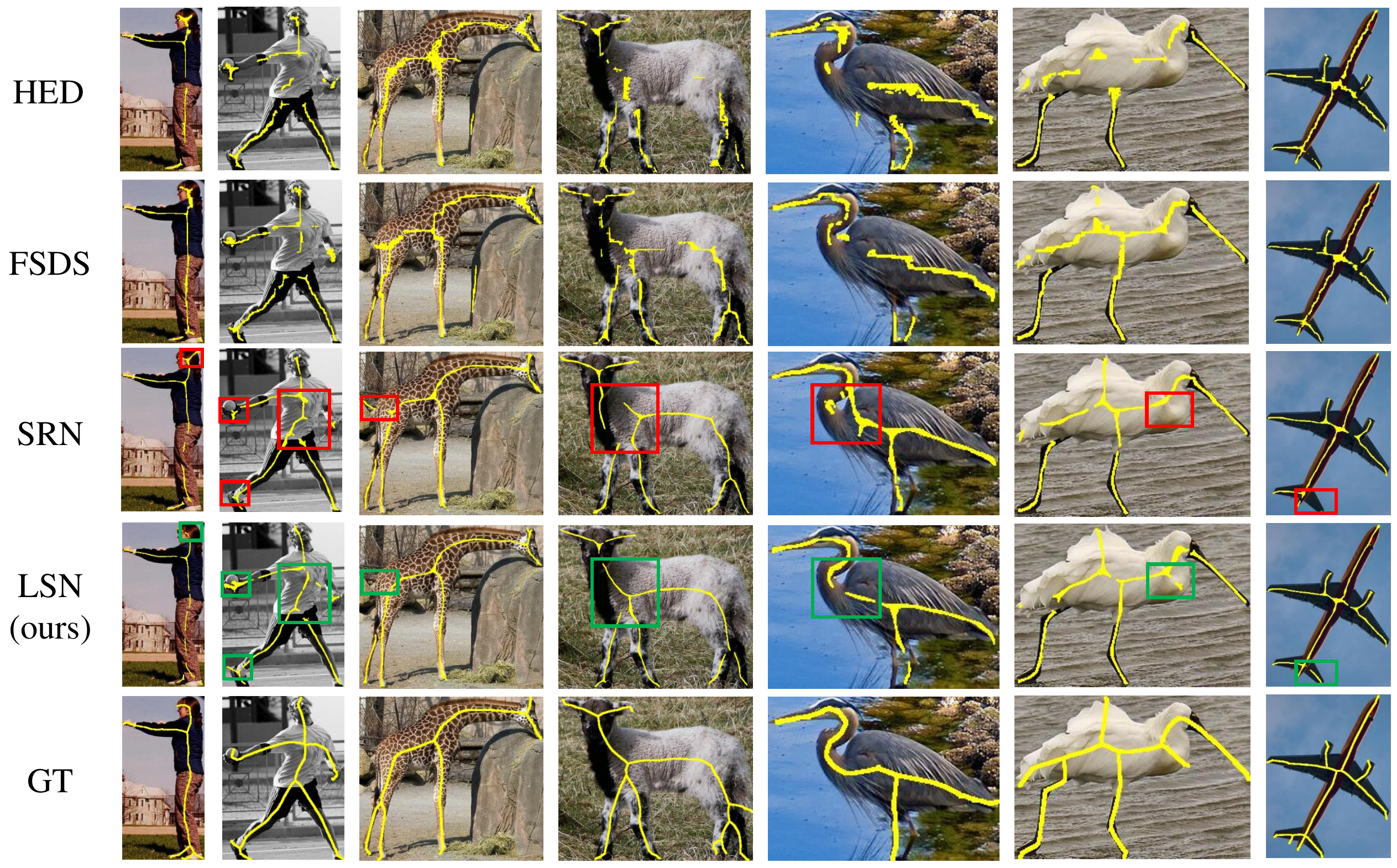}
\caption{Skeleton detection examples by state-of-the-art approaches including HED \cite{ref6}, FSDS \cite{ref7}, SRN \cite{ref1}, and LSN. The red boxes are false positive or dismiss in SRN, while the blue ones are correct reconstruction skeletons in LSN at the same position. (Best viewed in color with zoon-in.)}
\label{fig::skeleton_examples}
\end{figure}

The results show that feature linear span is efficient for skeleton detection. As discussed above, HED and SRN are two special case of LSN. LSN that used three spanned layers in each span unit is a better choice than the state-of-the art SRN. Some skeleton detection results are shown in Fig. \ref{fig::skeleton_examples}. It is illustrated that HED produces lots of noise while the FSDS is not smooth. Comparing SRN with LSN, one can see that LSN rectifies some false positives as shown in column one and column three and reconstruct the dismiss as shown in column six. 
\setlength{\tabcolsep}{4pt}
\begin{table}
\begin{center}
\caption{Performance comparison  of the state-of-the-art approaches on the public WH-SYMMAX \cite{ref17}, SK-SMALL \cite{ref7}, SYMMAX \cite{ref15}, and Sym-PASCAL \cite{ref1} datasets.}
\begin{tabular}{lcccc}
\hline\noalign{\smallskip}
 & WH-SYMMAX  &SK-SMALL  &	SYMMAX  &	Sym-PASCAL \\
\noalign{\smallskip}
\hline
\noalign{\smallskip}
Levinshtein \cite{ref12} &	0.174 &	0.217 &	-- &	0.134 \\
Lee \cite{ref13} &	0.223 &	0.252 &	-- &	0.135 \\
Lindeberg \cite{ref37} &	0.277 &	0.227 &	0.360 &	0.138 \\
Particle Filter \cite{ref14} &	0.334 &	0.226 &	-- &	0.129 \\
MIL \cite{ref15} &	0.365 &	0.392 &	0.362 &	0.174 \\
\hline
HED \cite{ref6} &	0.743 &	0.542 &	0.427 &	0.369 \\
FSDS \cite{ref7} &	0.769 &	0.623 &	0.467 &	0.418 \\
SRN \cite{ref1} &	0.780 &	0.632 &	0.446 &	\textbf{0.443} \\
LSN (ours) &	\textbf{0.797} &	\textbf{0.633} &	\textbf{0.480} &	0.425 \\
\hline
\label{tab::four_datasets}
\end{tabular}
\end{center}
\end{table}
\setlength{\tabcolsep}{1.4pt}


The proposed LSN is also evaluated on other four commonly used datasets, including WH-SYMMAX \cite{ref17}, SK-SMALL \cite{ref7}, SYMMAX \cite{ref15}, and Sym-PASCAL \cite{ref1}. The F-measure are shown in Table\ \ref{tab::four_datasets}. Similar with SK-LARGE, LSN achieves the best detection performance on WH-SYMMAX, SK-SMALL, and SYMMAX, with the F-measure 79.7\%, 63.3\% and 48.0\%. It achieves 5.4\%, 8.1\%, and 5.3\% performance gain compared with HED, and 1.7\%, 0.1\%, and 2.4\% gain compared with SRN. On Sym-PASCAL, LSN achieves comparable performance of 42.5\% vs. 44.3\% with the state-of-the art SRN.

\textbf{Edge detection.} 
Edge detection task has similar implementation with skeleton that discriminate whether a pixel belongs to an edge. It also can be reconstructed by the convolutional feature maps. In this section, we compare the edge detection result of the proposed LSN with some other state-of-the-art methods, such as Canny \cite{ref40}, Sketech Tokens \cite{ref39}, Structured Edge (SE) \cite{ref38}, gPb \cite{ref35}, DeepContour \cite{ref18}, HED \cite{ref6}, and SRN \cite{ref1}, Fig. 8 and Table 5.

In Fig. 8, it is illustrated that the best conventional approach is SE with F-measure (ODS) of 0.739 and all the CNN based approaches achieve much better detection performance. HED is one of the baseline deep learning method, which achieved 0.780. The proposed LSN reportes the highest F-measure of 0.790, which has a very small gap (0.01) to human performance. The F-measure with an optimal scale for the per image (OIS) was 0.806, which was even higher than human performance, Table 5. The good performance of the proposed LSN demonstrates its general applicability to image-to-mask tasks.

{
\setlength\parindent{-0.5em}
\vspace{1em}
\begin{minipage}{.45\textwidth}
\centering
\includegraphics[width=2.2in]{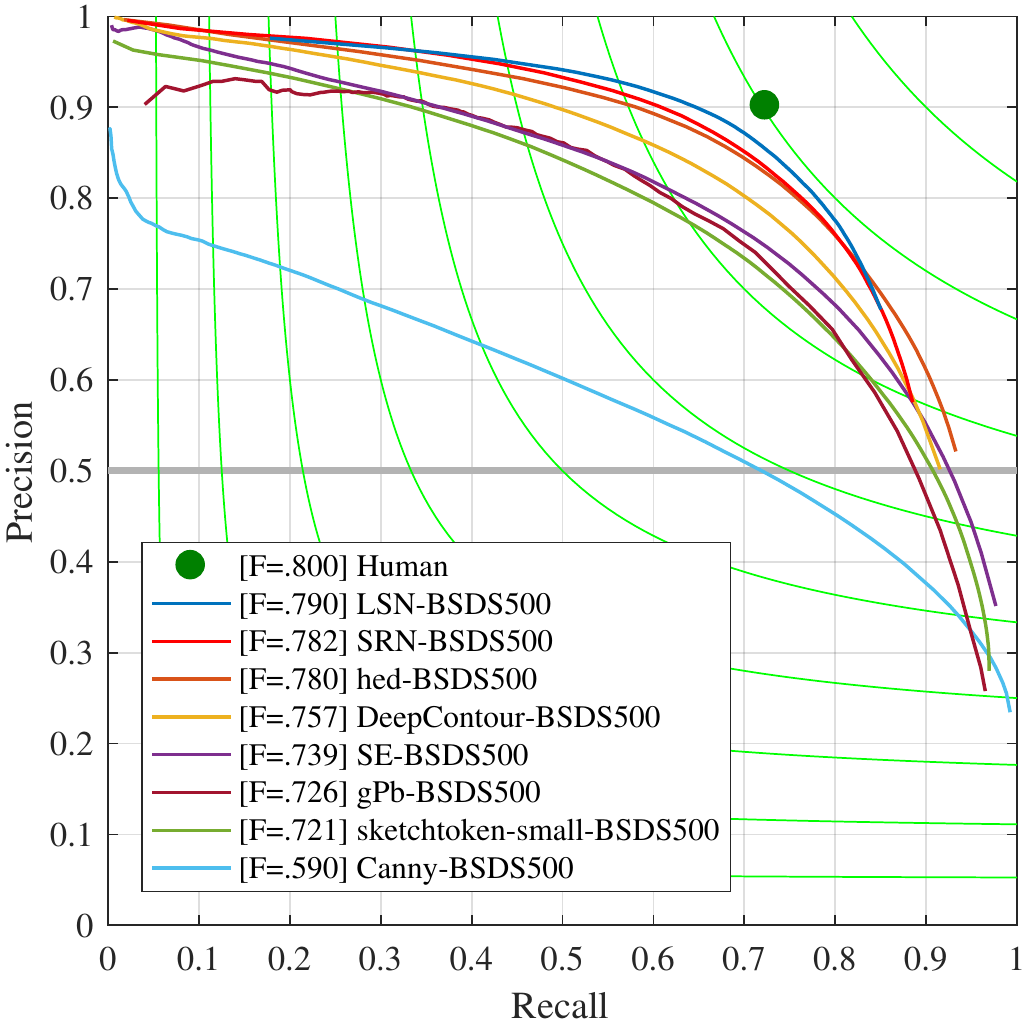} 
\makeatletter\def\@captype{figure}\makeatother
\caption{The PR-curve on the BSDS500 edge deteciton dataset.}
\end{minipage} \quad
\begin{minipage}{.52\textwidth}
  \makeatletter\def\@captype{table}\makeatother
  \caption{Performance comparison on the BSDS500 edge detection dataset. $\dagger$ GPU time.}
  \begin{tabular}{lcccc}
    \hline\noalign{\smallskip}
    Mehods & ODS & OIS & AP & FPS \\
    \noalign{\smallskip}
    \hline
    \noalign{\smallskip}
    Canny \cite{ref40} &	0.590 &	0.620 &	00578 &	15 \\
    ST \cite{ref39} & 0.721 &	0.739 &	0.768 &	1 \\
    gPb \cite{ref35} & 0.726 & 0.760 & 0.727 & 1/240 \\
    SE \cite{ref18} &	0.739 &	0.759 &	0.792 &	2.5 \\
    \hline
    DC \cite{ref18} &	0.757 &	0.776 &	0.790 &	1/30 $\dagger$ \\
    HED \cite{ref6} &	0.780 &	0.797 &	0.814 &	2.5 $\dagger$ \\
    SRN \cite{ref1} &	0.782 &	0.800 &	0.779 &	2.3 $\dagger$ \\
    LSN (ours) &	\textbf{0.790} & \textbf{0.806} &	0.618 &	2.0 $\dagger$ \\
    \hline
    Human &	0.800 &	0.800 &	-- & -- \\	
    \hline
 \end{tabular}    
  \label{fig:sklarge}
\end{minipage}
\vspace{1em}
}

\section{Conclusion}

Skeleton is one of the most representative visual properties, which describes objects with compact but informative curves. In this paper, the skeleton detection problem is formulated as a linear reconstruction problem. Consequently, a generalized linear span framework for skeleton detection has been presented with formal mathematical definition. We explore the Linear Span Units (LSUs) to learn a CNN based mask reconstruction model. With LSUs we implement three components including feature linear span, resolution alignment, and subspace linear span, and update the Holistically-nested Edge Detection (HED) network to Linear Span Network (LSN). With feature linear span, the ground truth space can be approximated by the linear spanned output space. With subspace linear span, not only the independence among spanning sets of subspaces can be increased, but also the spanned output space can be enlarged. As a result, LSN will have better capability to approximate the ground truth space, $i.e.$, against the cluttered background and scales. Experimental results validate the state-of-the-art performance of the proposed LSN, while we provide a principled way to learn more representative convolutional features.
\section*{Acknowledgement}
This work was partially supported by the National Nature Science Foundation of China under Grant 61671427 and Grant 61771447, and Beijing Municipal Science \& Technology Commission under Grant Z181100008918014.
\bibliographystyle{splncs04}
\bibliography{1576}
\end{document}